\documentclass[runningheads]{llncs}
\usepackage{graphicx}
\usepackage{booktabs}
\usepackage{multirow}
\usepackage{url}
\usepackage[super]{nth}
\usepackage{color, colortbl}
\usepackage[colorlinks=true, allcolors=blue]{hyperref}
\usepackage{orcidlink}
\usepackage{fontawesome}
\usepackage{amsmath} 
\usepackage{amsfonts} 

\begin{document}

\title{A Federated Learning-Friendly Approach for Parameter-Efficient Fine-Tuning of SAM in 3D~Segmentation}

%
\titlerunning{FLAP-SAM}
\author{Mothilal Asokan   \orcidlink{0009-0005-9461-2749} \and 
Joseph Geo Benjamin\orcidlink{0009-0001-4680-9134}\textsuperscript{(\faEnvelopeO)} \and 
Mohammad Yaqub    \orcidlink{0000-0001-6896-1105} \and 
Karthik Nandakumar\orcidlink{0000-0002-6274-9725} 
}
\authorrunning{M. Asokan \textit{et al.}}

\institute{
Mohamed bin Zayed University of Artificial Intelligence (MBZUAI), \\ Abu Dhabi, United Arab Emirates \\
\email{\{mothilal.asokan, joseph.benjamin, mohammad.yaqub, karthik.nandakumar\}@mbzuai.ac.ae} \\
}

\maketitle              

%
\begin{abstract}
Adapting foundation models for medical image analysis requires finetuning them on a considerable amount of data because of extreme distribution shifts between natural (source) data used for pretraining and medical (target) data. 
However, collecting task-specific medical data for such finetuning at a central location raises many privacy concerns. 
Although Federated learning (FL) provides an effective means for training on private decentralized data, communication costs in federating large foundation models can quickly become a significant bottleneck, impacting the solution's scalability.
In this work, we address this problem of `efficient communication while ensuring effective learning in FL' by combining the strengths of Parameter-Efficient Fine-tuning (PEFT) with FL. 
Specifically, we study plug-and-play Low-Rank Adapters (LoRA) in a federated manner to adapt the Segment Anything Model (SAM) for 3D medical image segmentation. Unlike prior works that utilize LoRA and finetune the entire decoder, we critically analyze the contribution of each granular component of SAM on finetuning performance. Thus, we identify specific layers to be federated that are very efficient in terms of communication cost while producing on-par accuracy. 
Our experiments show that retaining the parameters of the SAM model (including most of the decoder) in their original state during adaptation is beneficial because fine-tuning on small datasets tends to distort the inherent capabilities of the underlying foundation model. 
On Fed-KiTS, our approach decreases communication cost ($\sim$$48\times\downarrow$) compared to full fine-tuning while increasing performance ($\sim$$6\%\uparrow$ Dice score) in 3D segmentation tasks. Our approach performs similar to SAMed while achieving $\sim$$2.8\times$ reduction in communication and parameters to be finetuned.  
We further validate our approach with experiments on Fed-IXI and Prostate MRI datasets.
Our code is available at \url{https://github.com/BioMedIA-MBZUAI/FLAP-SAM}.
 
\end{abstract}

%

\section{Introduction}

Segmentation is one of the cornerstone tasks in modern medical image analysis for automated diagnosis and disease monitoring. While the advent of foundational models has pushed the boundaries of the state-of-the-art in many computer vision applications, such benefits are yet to transfer fully to the medical imaging domain \cite{foundsurvey-ZHANG2024102996}. For example, the Segment Anything Model (SAM) \cite{sam0-Kirillov_2023_ICCV} has an excellent zero-shot generalization to new distributions and tasks involving natural images. However, the SAM model fails to generalize well across diverse medical imaging modalities due to the insurmountable distribution shifts \cite{samm-huix2024natural_transferability}. Works like \cite{samm-he2023accuracy,samm-huang2024segment} highlight a substantial performance gap between the zero-shot inference and training on domain-specific medical images despite using various prompts in SAM.
MSA \cite{wu2023medical} and SAM-Med2D \cite{samm-cheng2023sam_promt} improve SAM by using tailored prompting techniques in 2D medical images. However, creating such prompts for each 2D slice of 3D data is labor-intensive.

\begin{figure}[t]
    \centering
    \includegraphics[width=0.45\linewidth]{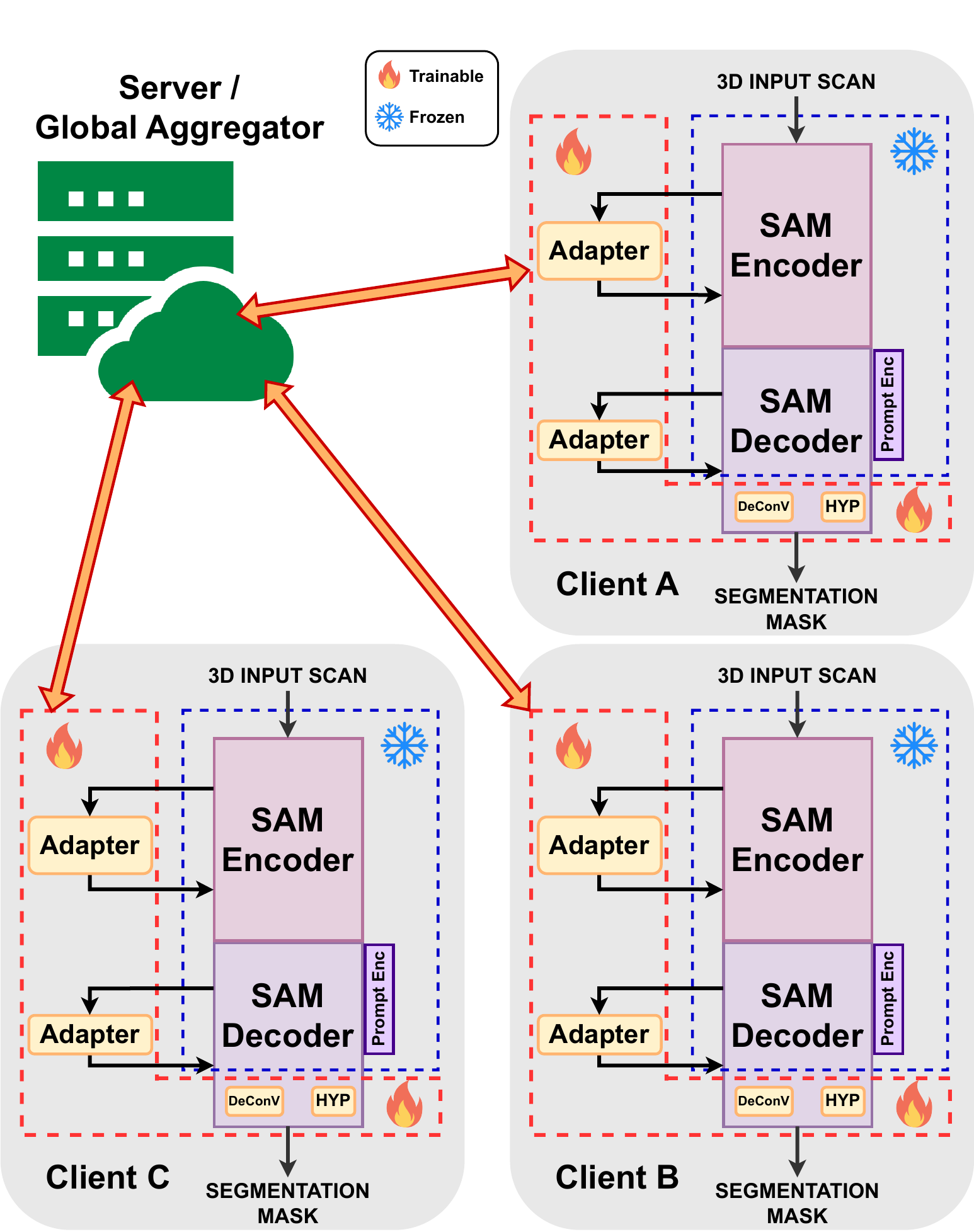}
    \hspace{1ex}
    \includegraphics[width=0.48\linewidth]{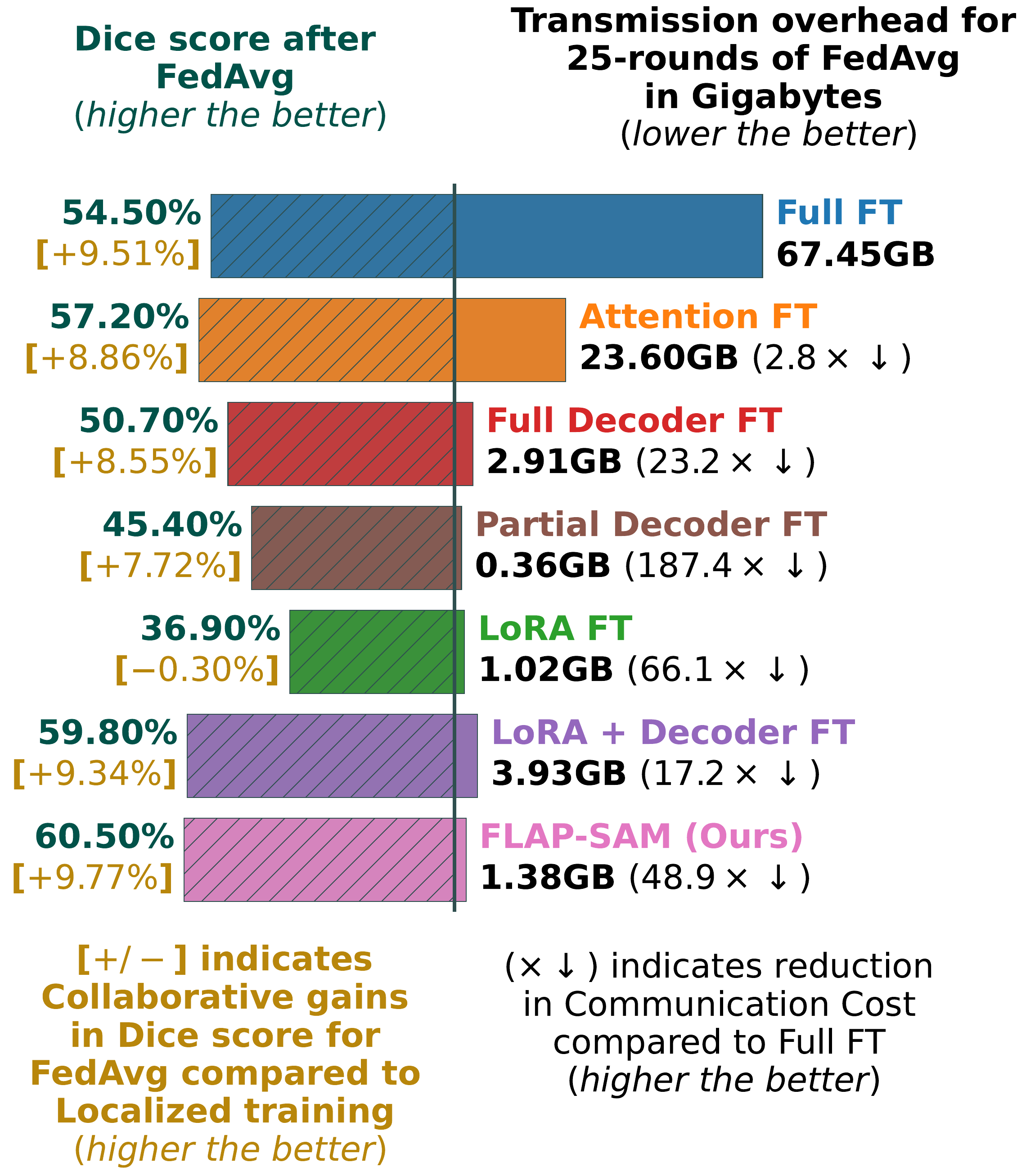}
    \caption{(Left) The proposed FLAP-SAM framework and (Right). The comparison (in terms of Dice score(\%), Collaborative gains(\%) and Communication Cost($\times \downarrow$)) of our method against other fine-tuning methods on Fed-KITS2019\cite{NEURIPS2022_232eee8e}.}
    \label{fig:intro-diagrams}
\end{figure}

The usual approach has been fine-tuning SAM for the target application using large task-specific datasets e.g., MedSAM\cite{samm-ma2024Medsam}. For tasks with significant distribution shifts with limited data, the decoder block of SAM is fine-tuned while leaving the encoder untouched. 
In contrast to fine-tuning all parameters, parameter-efficient fine-tuning(PEFT) methods fine-tune only a minimal number of parameters using representative data such as prompt tuning \cite{lora2-jia2022visual} and low-rank adapters (LoRA) \cite{lora0-hu2021lora}. Several works \cite{sam1-paranjape2023adaptivesam_promt,wang2023sam,sam0-zhang2023customizedSam} have employed LoRA for fine-tuning, resulting in superior performance in various 2D segmentation tasks. For 3d segmentation, \cite{sam0-chen2023MAsam} uses factor tuning (FacT) adapters \cite{jie2023fact} for training.
However, accessing diverse medical data for such finetuning is not always feasible because relevant datasets specific to a medical task may not be available with any single entity. It is often spread across multiple institutions and cannot be shared due to privacy and confidentiality constraints \cite{NEURIPS2022_232eee8e}. 
While the data decentralization issue can be handled effectively with federated learning (FL) \cite{fed-fedaverage,fed-li2020federated}, the ginormous parameter size of SAM ($\sim$100M-700M) makes it impractical for FL as it imposes a substantial communication cost.

In this work, we uncover the importance of fine-tuning various components of SAM when adapted for 3D medical segmentation and propose \textbf{FLAP-SAM}, \textit{a PEFT approach involving LoRA that is amenable to FL} (see Fig~\ref{fig:intro-diagrams}). Moreover, high parameter efficiency reduces communication costs in FL and prevents overfitting in data-limited scenarios. With methods like MA-SAM\cite{sam0-chen2023MAsam}, which uses FacT adapters for 3D segmentation, it is challenging to federate because of the tensor decomposition involved. Hence, we employ LoRA, which is FL-friendly, to customize SAM for 3D medical image segmentation.
SAMed\cite{sam0-zhang2023customizedSam} uses a similar approach where LoRA and the entire SAM decoder are fine-tuned for 2D segmentation. Our approach selectively finetunes certain decoder parts, reducing parameters and communication costs.

\section{Preliminaries}

\noindent \textbf{Overview of SAM:}
The architecture of SAM\cite{sam0-Kirillov_2023_ICCV} can be decoupled into three major components: the \textit{Image Encoder} (\texttt{IE}) to compute image embeddings, the \textit{Prompt Encoder} (\texttt{PE}) to generate prompt embeddings, and the \textit{Mask Decoder} (\texttt{MD}) that combines the image and prompt embeddings to generate segmentation masks as shown in Fig~\ref{fig:methods-diagram}. 
Utilizing ViT \cite{vit-dosovitskiy2020image} as the backbone, \texttt{IE} extracts image features through a sequence of $L$ transformer blocks. Meanwhile, \texttt{PE} takes various input prompts in the form of points, boxes, or masks and encodes them into prompt embeddings to aid in segmentation tasks. 
We operate SAM in the \textit{fully automatic} mode, in which a regular grid of foreground points is presented as input prompts to the \texttt{PE}, thus eliminating the dependence on user-defined prompts. \texttt{MD} performs cross-attention between the image and prompt embeddings, employing transposed convolutional layers for up-scaling back to image dimension (\texttt{UP}) and a hyper multi-layer perceptron (\texttt{HYP}) to produce segmentation masks. 
Following \cite{sam0-chen2023MAsam}, we use a slightly modified SAM mask decoder that has two additional transposed convolutional layers, which up-sample the feature maps by $16 \times$ to match the resolution of the input while ensuring improved discrimination of small anatomical structures or lesions in medical images~\cite{ronneberger2015u}.

For simplicity, let $\theta_{\texttt{IE}}$, $\theta_{\texttt{PE}}$, and $\theta_{\texttt{MD}}$ denote the parameters of $\texttt{IE}$, $\texttt{PE}$, and $\texttt{MD}$, respectively. Also, $\theta_{\texttt{IE}}$ can be further partitioned as $\theta_{\texttt{IE-AT}}$ and $\theta_{\texttt{IE-NA}}$, where $\theta_{\texttt{IE-AT}}$ denotes the parameters of all the attention layers within $\texttt{IE}$ and $\theta_{\texttt{IE-NA}}$ represents all the other parameters in \texttt{IE} not related to the attention layers. On the other hand, $\theta_{\texttt{MD}}$ can be partitioned as $\theta_{\texttt{MD-TR}}$, $\theta_{\texttt{MD-UP}}$, and $\theta_{\texttt{MD-HYP}}$, where $\theta_{\texttt{MD-TR}}$ denotes the parameters of all the transformer blocks within $\texttt{MD}$ (such as self-attention, cross-attention from tokens to image embeddings (t2i), and cross-attention from image embeddings to tokens (i2t)), $\theta_{\texttt{MD-UP}}$ denotes the parameters of the transposed convolutional layers used for upscaling, and $\theta_{\texttt{MD-HYP}}$ represents the parameters of \texttt{HYP}. 

\noindent \textbf{PEFT Formulation:}
The most straightforward approach to adapt a SAM model for a downstream task is to fine-tune all its parameters, including $\theta_{\texttt{IE}}$, $\theta_{\texttt{PE}}$, and $\theta_{\texttt{MD}}$. This full fine-tuning (FullFT) strategy requires more memory footprint to store a copy of all the updated parameters and often leads to overfitting when the data is severely limited. Recent works have shown that fine-tuning only the attention layers of a transformer encoder is sufficient for good adaptation \cite{vit-touvron2022three}. This approach is called attention fine-tuning (AttnFT), where only $\theta_{\texttt{IE-AT}}$ and $\theta_{\texttt{MD}}$ are updated. 
Typically, the attention-related parameters of a transformer encoder constitute one-third of its overall parameters. Thus, AttnFT leads to some improvement in parameter efficiency and generally provides good performance on downstream tasks. 
Another common approach is to freeze the image encoder and fine-tune the entire mask decoder. This is because the cross-attention layers in the decoder focus on specific patches in the image embeddings corresponding to the prompts and transform them into segmentation predictions \cite{samm-xie2024sam_fewshotFT}. We call this approach DecFT, where $\theta_{\texttt{MD}}$ is updated.
It is also possible to freeze all the parameters and fine-tune only the output layers, namely, \texttt{UP} and \texttt{HYP}. This approach is analogous to linear probing in classification tasks, which we refer to as partial decoder fine tuning (PDecFT). Since only a small fraction of parameters, namely, $\theta_{\texttt{MD-UP}}$, and $\theta_{\texttt{MD-HYP}}$ are updated, PDecFT has high parameter efficiency but usually provides only sub-optimal performance.

\noindent \textbf{LoRA adapters:}
Low-rank adaptation \cite{lora0-hu2021lora} is a promising PEFT technique widely used for adapting foundation models to downstream tasks. Each attention layer within a transformer block has four weight (projection) matrices $W_q$ (query), $W_k$ (key), $W_v$ (value), and $W_o$ (output), where each $W \in \mathbb{R}^{d \times d'}$. The core idea of LoRA is to constrain the modifications to a pre-trained weight matrix $W$ to a linear update matrix $\Delta W$, which can be further constrained using a low-rank decomposition, i.e., $\Delta W = BA$, where $B \in \mathbb{R}^{d \times r}, A \in \mathbb{R}^{r \times d'}$ and the rank $r \ll \min\{d, d'\}$. This approach effectively reduces the parameter space while preserving the essential information needed for adaptation. $W$ is frozen during fine-tuning, and only $A$ and $B$ matrices are updated. For input $x$, the output $\tilde{x}$ is computed as follows:
\begin{equation}\label{eq:lora}
    \tilde{x} = (W + \alpha \Delta W)x =  Wx + \alpha \Delta Wx = Wx + \alpha BAx, 
\end{equation}

\noindent where $\alpha$ is a scale parameter. Following \cite{lora0-hu2021lora}, we employ LoRA only to the projection matrices $W_q$ and $W_v$ in all the $\Gamma$ attention layers of \texttt{IE} and \texttt{MD}. Let $\theta_{\texttt{LoRA}}$ represent the set of all LoRA parameters, where $\theta_{\texttt{LoRA}} = \{A^{\ell}_{q}, A^{\ell}_{v}, B^{\ell}_{q}, B^{\ell}_{v}\}^{\Gamma}_{\ell=1}$. When only the LoRA parameters are updated during fine-tuning, we refer to this case as LoRAFT. While this approach also has good parameter efficiency, it typically results in sub-optimal performance for segmentation tasks.
To improve this, in SAMed\cite{sam0-zhang2023customizedSam}, both the decoder and LoRA are fine-tuned. Here, $\theta_{\texttt{MD}}$ and $\theta_{\texttt{LoRA}}$ parameters are updated together, and we represent this approach as LoRADecFT.

\noindent \textbf{Federated Learning:}
In an FL system, \textit{K} clients can collaboratively train a global model with parameters $\Theta$. 
The goal is to solve the following optimization problem:
\begin{equation}
    \min_{\Theta \in \mathbb{R}^T} \frac{1}{K} \sum_{k=1}^{K} \mathcal{L}^{seg}_k(\Theta),
    \label{eq:fedavgloss}
\end{equation}

\noindent where $\mathcal{L}^{seg}_k(\Theta) = E_{x \sim \mathcal{D}_k}[\mathcal{L}^{seg}_k(\Theta; x)]$ is the loss function of the $k^{th}$ client ($k \in [1,K])$, $\mathcal{D}_k$ represents the data distribution of the $k^{th}$ client, and $T = |\Theta|$ is the number of parameters that need to be learned. Note that if the distributions $\mathcal{D}_i$ and $\mathcal{D}_j$ are different for clients $i$ and $j$, the scenario is referred to as non-iid (not independent and identically distributed). A widely used method for solving the optimization problem is FedAvg \cite{fed-fedaverage}.  At each round, the server broadcasts the global model to each client. Then, all clients conduct local training on their data and send back the updated model to the server. Finally, the server updates the global model as a weighted average of these local model updates. The server update at round $n$ in FedAvg can be formulated as follows:
\begin{equation}
    \Theta^{n+1} = \sum_{k=1}^{K} \alpha_k \Theta^{n}_{k},
    \label{eq:fedavg}
\end{equation}
\noindent where $\Theta^{n}_{k}$ denotes the local model of $k^{\text{th}}$ client in round $n$ and $\alpha_k$ is the weight assigned for each client. 

\section{Proposed FLAP-SAM Approach}

\begin{figure}[t]
    \centering
    \includegraphics[width=0.95\linewidth]{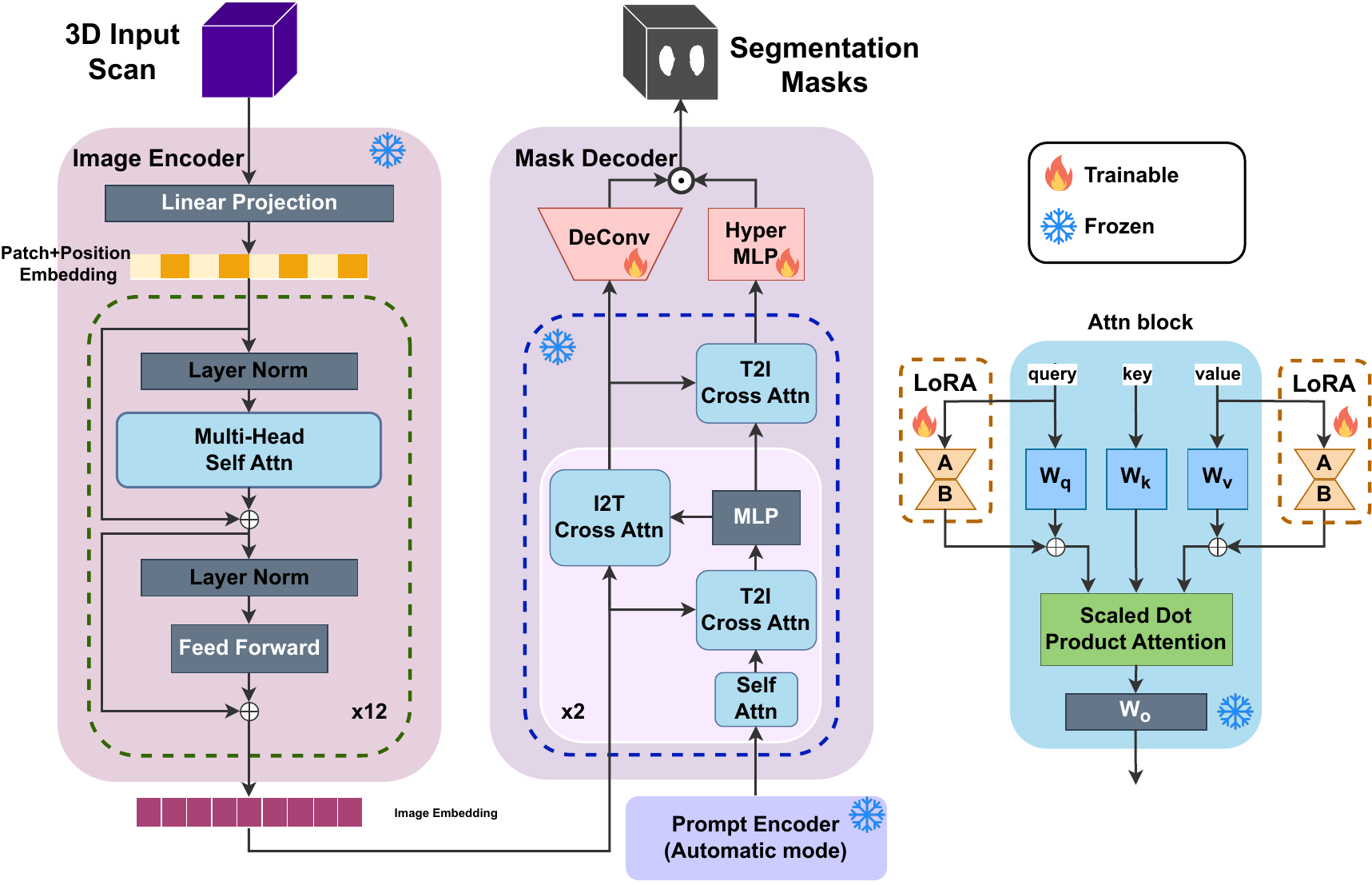}
    \caption{Architecture of the proposed plug-and-play adapter. Only the decoder output layers and LoRAs are fine-tuned, while the remaining parameters in SAM are frozen.}
    \label{fig:methods-diagram}
\end{figure}

We aim to efficiently and effectively adapt the SAM for medical image segmentation tasks using limited data distributed across multiple entities. 
Based on this consideration, we propose to update both $\theta_{\texttt{LoRA}}$ as well as the final decoder output layers $\theta_{\texttt{MD-UP}}$ and $\theta_{\texttt{MD-HYP}}$. This leads to the proposed fine-tuning for SAM, a hybrid of PDecFT and LoRAFT, as shown in Fig~\ref{fig:methods-diagram}. Though there is a marginal increase in the number of parameters compared to LoRAFT, the proposed approach performs well because FLAP-SAM provides enough flexibility to be effectively fine-tuned. Moreover, since almost all the parameters of the original foundation model are retained without modification, its inherent capabilities remain unaffected. Another benefit is its memory efficiency, and the small parameter size of the proposed adapter makes it possible to learn them collaboratively via FL while greatly reducing communication costs.

When aggregating LoRA parameters $(\theta_{\texttt{LoRA}})$, the $k^{th}$ client sends 
$\{A^{\ell}_{q,k}, A^{\ell}_{v,k}$, $B^{\ell}_{q,k}, B^{\ell}_{v,k}\}^{\Gamma}_{\ell=1}$ to the server. 
The server first needs to reconstruct $\Delta W^{\ell}_{q,k} = B^{\ell}_{q,k}\cdot A^{\ell}_{q,k}$ and $\Delta W^{\ell}_{v,k} = B^{\ell}_{v,k}\cdot A^{\ell}_{v,k}$ for each $\ell$ and $k$, then performs FedAvg as shown in Eq. \eqref{eq:fedavg} to get the aggregated global weight matrices $\Delta W^{\ell}_q$ and $\Delta W^{\ell}_v$ of each attention layer $\ell$. 
Finally, the server applies singular value decomposition to decompose the aggregated matrices back to global LoRA parameters $\{A^{\ell}_{q}, B^{\ell}_{q}, A^{\ell}_{v}$, $B^{\ell}_{v}\}^{\Gamma}_{\ell = 1}$, which are sent back to the clients. We refer to this federated learning of plug-and-play SAM adapter as FLAP-SAM.


\definecolor{BaseColor}{rgb}{0.717, 0.913, 0.968}
\definecolor{OurColor}{rgb}{0.984, 0.843, 0.929}


\begin{table}[!ht]
\centering
\caption{Comparison on all datasets for different fine-tuning methods. 
$\ddag -$ FL setting of MA-SAM is not feasible since decomposing FacT tensors after aggregation is not possible; a centralized score is provided for performance comparison in 3D segmentation. 
$** -$ Parameter counts for single class segmentation task (add $0.134M$ params for each additional class).  
The baseline (full fine-tuning) is highlighted in \colorbox{BaseColor}{Blue} and our method in \colorbox{OurColor}{Pink}. 
}

\label{tab:full_results_table}
{\renewcommand{\arraystretch}{1.1}
\begin{tabular}{c rc cc cc cc cc}
\toprule
\midrule
\multirow{2}{*}{\textbf{Experiments}} & \multirow{2}{*}{\textbf{Setting}} &  \multicolumn{3}{c}{\textbf{Mean Dice score}} & & \multirow{2}{*}{\begin{tabular}[c]{@{}c@{}}\textbf{Trainable}/ \\ \textbf{Total params}$^{**}$\end{tabular}} \\ \cline{3-5} 
\multicolumn{1}{l}{} && \textbf{Fed-KiTS} & \textbf{Fed-IXI} & \textbf{Prost.MRI}  & \\ \hline
\rowcolor{BaseColor} \cellcolor{BaseColor} 
& Local & 0.4493 & 0.9777 & 0.8421 && \\
\rowcolor{BaseColor} \cellcolor{BaseColor} 
& Federated & 0.5444 & 0.9811 & 0.9084 && 90.399M/90.399M \\
\rowcolor{BaseColor} \cellcolor{BaseColor} \multirow{-3}{*}{\begin{tabular}[c]{@{}c@{}}\textbf{FullFT}\\ (baseline) \\ \{$\theta_{\texttt{IE}}, \theta_{\texttt{PE}}, \theta_{\texttt{MD}}$\} \end{tabular}} 
& Centralized & 0.5274 & 0.9834 & 0.8955 && (100\%) \\ \hline

\multirow{3}{*}{\begin{tabular}[c]{@{}c@{}} \textbf{AttnFT}\\ \{$\theta_{\texttt{IE-AT}}, \theta_{\texttt{MD}}$\} \end{tabular}}
 & Local & 0.4838 & 0.8848 & 0.6315 && \\
 & Federated & 0.5724 & 0.9674 & 0.8797 && 29.575M/90.399M \\
 & Centralized & 0.5486 & 0.9774 & 0.8957 && (32.7\%) \\ \hline

\multirow{3}{*}{\begin{tabular}[c]{@{}c@{}} \textbf{DecFT}\\ \{$\theta_{\texttt{MD}}$\} \end{tabular}}
 & Local & 0.4213 & 0.9750 & 0.8200 && \\
 & Federated & 0.5068 & 0.9771 & 0.8101 && 3.768M/90.399M\\
 & Centralized & 0.5179 & 0.9789 & 0.8587 && (4.2\%) \\ \hline

\multirow{3}{*}{\begin{tabular}[c]{@{}c@{}} \textbf{LoRAFT}\\ \{$\theta_{\texttt{LoRA}}$\} \end{tabular}}
 & Local & 0.3717 & 0.9728 & 0.8386 && \\
 & Federated & 0.3687 & 0.9777 & 0.8578 && 1.368M/91.767M \\
 & Centralized & 0.5242 & 0.9798 & 0.8893 && (1.5\%) \\ \hline

\multirow{3}{*}{\begin{tabular}[c]{@{}c@{}} \textbf{LoRADecFT} \\ (SAMed)\cite{samm-ma2024Medsam}\\ \{$\theta_{\texttt{LoRA}}, \theta_{\texttt{MD}}$\} \end{tabular}}
 & Local       & 0.5053 & 0.9829 & 0.8929 && \\
 & Federated   & 0.5987 & 0.9836 & 0.8949 && 5.270M/91.767M \\
 & Centralized & 0.6100 & 0.9852 & 0.9039 && (5.8\%) \\ \hline

\multirow{3}{*}{\begin{tabular}[c]{@{}c@{}} \textbf{PDecFT}\\ \{$\theta_{\texttt{UP}}, \theta_{\texttt{HYP}}$\} \end{tabular}}
 & Local & 0.3764 & 0.9678 & 0.7890 && \\
 & Federated & 0.4536 & 0.9693 & 0.7017 && 0.344M/90.399M \\
 & Centralized & 0.4793 & 0.9711 & 0.8008 && (0.4\%) \\ \hline

\rowcolor{OurColor} \cellcolor{OurColor}
& Local & 0.5069 & 0.9829 & 0.8845 && \\
\rowcolor{OurColor} \cellcolor{OurColor} 
& Federated & 0.6046 & 0.9834 & 0.8867 && 1.712M/91.767M \\
\rowcolor{OurColor} \cellcolor{OurColor} \multirow{-3}{*}{\begin{tabular}[c]{@{}c@{}} \textbf{FLAP-SAM}\\ \textbf{(ours)} \\ \{$\theta_{\texttt{LoRA}}, \theta_{\texttt{UP}}, \theta_{\texttt{HYP}}$\}\end{tabular}} 
& Centralized & 0.5980 & 0.9851 & 0.9044 && (1.9\%) \\ 
\midrule

\multirow{1}{*}{\begin{tabular}[c]{@{}c@{}} \textbf{MA-SAM$^\ddag$}\cite{sam0-chen2023MAsam}\end{tabular}}
 & Centralized & 0.6023 & 0.9707 & 0.9125 && {\begin{tabular}[c]{@{}c@{}} 28.667M/115.298M \\(25\%)\end{tabular}}\\ 
\midrule
\bottomrule
\end{tabular}
}
\end{table}

\section{Experiments}
\noindent\textbf{Datasets}: We utilize \textbf{\textit{Fed-KITS2019}}, a 6-client federated version of the KiTS19 dataset from FLamby \cite{NEURIPS2022_232eee8e}, which was created from the Kidney Tumor Segmentation Challenge 2019 in CT scans \cite{heller2020state,heller2019kits19}.
Each client's train/test split is 9/3, 11/3, 9/3, 9/3, 12/4, and 24/6.
The preprocessing pipeline comprises intensity clipping (\nth{5} and \nth{95} percentile of image intensities of each client were calculated) followed by z-scale normalization, where we subtract the mean and divide by the standard deviation of the image intensities. 
\textbf{\textit{Fed-IXI}}, extracted from the Information eXtraction from Images - IXI database \cite{perez2021torchio,BrainDevelopment} of brain T1 MRIs from 3 hospitals (Guys, HH, and IOP) contains $249/62, 145/36$ and $59/15$ train/test splits respectively.
In a preprocessing step, min-max normalization was applied to each scan and padded with zeros in the axial plane (final shape $83\times64\times64$). 
\textbf{\textit{Prostate MRI}} is a multi-site segmentation dataset proposed by Liu \textit{et al.} \cite{liu2020saml}, comprises prostate T2-weighted MRI data from six different data sources (\textit{i.e.,} Site A to F) out of the three publicly available datasets: NCI-ISBI13 dataset \cite{bloch2015nci}, I2CVB dataset \cite{LEMAITRE20158} and PROMISE12 dataset \cite{LITJENS2014359}.
Each site has $30,30,19,13,12,12$ MRI scans of patients respectively and were randomly divided into train($\approx80\%$) and test($\approx20\%$) sets.
Since they were acquired with varying imaging protocols and contain heterogeneous data distributions, we normalized each site to zero mean and unit variance to reduce the intensity variance among different sites. We resized it to $224\times224$ in the axial plane.


\noindent\textbf{Implementation details}: We follow the input format and data augmentation as described in \cite{sam0-chen2023MAsam}, conducting all experiments using the  ``\verb|vit_b|" version of SAM on an NVIDIA A100-SXM4-40GB GPU.
The input to the model is of size ($N\times H\times W$), which consists of every set of $N$ consecutive slices ($N=5$). 
In LoRA, we initialize matrix $A$ from a random Gaussian distribution while setting matrix $B$ to zero and rank to $32$. 
The fine-tuning process employs a hybrid segmentation loss, combining cross-entropy loss and Dice loss as $\mathcal{L}^{seg}=\alpha\mathcal{L}^{CE} + \beta\mathcal{L}^{Dice}$, with weighting factors $\alpha=0.2$ and $\beta=0.8$ following \cite{sam0-zhang2023customizedSam}.
The training utilizes the Adam optimizer with a batch size of 32. 
We compare federated learning with localized (using client-owned data alone) and centralized (all data pooled) settings. We test each site data separately for all fine-tuning strategies, and the results are tabulated in Table~\ref{tab:full_results_table}.

\begin{table}[!ht]
\centering
\caption{Average Dice score across local test data in the federated setting on Fed-KiTS19 dataset. (Left) different rank values of LoRA; (Right) different low-rank adapter methods.}
\label{tab:merged_ablations}
{ \renewcommand{\arraystretch}{1.1}
\begin{tabular}{cc cc cc}
\toprule
\textbf{\begin{tabular}[c]{@{}c@{}}LoRA \\ rank\end{tabular}} && \textbf{\begin{tabular}[c]{@{}c@{}}Mean \\ Dice\end{tabular}} && \textbf{\begin{tabular}[c]{@{}c@{}}Trainable /\\ Total params\end{tabular}} &\\ \hline
\textbf{32} & & \textbf{0.605} & & 1.846M/91.901M & \\
\textbf{16} & & 0.599 & & 1.162M/91.217M & \\
\textbf{4}  & & 0.600 & & 0.649M/90.704M & \\ 
\bottomrule
\end{tabular}
\hspace{1.5em}
\begin{tabular}{cc cc cc}
\toprule
\textbf{Adapter} && \textbf{\begin{tabular}[c]{@{}c@{}}Mean \\ Dice\end{tabular}} &&
\textbf{\begin{tabular}[c]{@{}c@{}}Trainable /\\ Total params\end{tabular}} &\\ \hline
\textbf{LoRA} & & \textbf{0.605} & & 1.846M/91.901M & \\
\textbf{DoRA} & & 0.592 & & 1.846M/91.901M & \\
\textbf{MoLE} & & 0.603 & & 4.583M/94.637M & \\ 
\bottomrule
\end{tabular}
}
\end{table}

\subsection{Results and Discussion}
Our proposed FLAP-SAM method achieves $6\%$ absolute improvement in Dice score compared to the FullFT approach, with a $\sim$$49\times$ reduction in communication overhead on Fed-KITS. Due to the small size of the dataset, the FullFT approach easily results in overfitting, highlighting the importance of using PEFT methods in limited data settings \cite{chen2022revisiting}. 
The Attention fine-tuning (AttnFT) only achieves half of our improvement ($2.8\%$ less than FLAP-SAM) and still incurs $\sim$$17\times$ more communication cost than our method. Both LoRAFT and PDecFT are more efficient but have lower Dice scores than our method.
The LoRADecFT achieves an equivalent dice score to our method, but our method is $\sim$$2.8\times$ more efficient regarding parameters and communication.
We conduct ablation on the rank parameter of LoRA in FL, and the results are shown in Table~\ref{tab:merged_ablations}. We observe that a lower LoRA rank significantly reduces the trainable parameters with a marginal degradation in the Dice score. We also conduct experiments with other low-rank adapters like DoRA\cite{lora-liu2024DoRA} and MoLE\cite{lora-wu2023MoLE}, which only show marginal performance differences among the adapters.

We also benchmark FLAP-SAM against MA-SAM \cite{sam0-chen2023MAsam}, which uses a 3D adapter along with FacT for fine-tuning. We perform this comparison only in the centralized setting because the decomposition of $\Delta W$ back to FacT-Tensor-Train or FacT-Tucker formats \cite{jie2023fact} after federated aggregation is not straightforward. Although MA-SAM produces comparable results to our method (see Table \ref{tab:full_results_table}) in a centralized setting, it uses $28.7$M trainable parameters ($\sim$$16\times$ more than our method).  
This validates our choice of using LoRA, which is both parameter-efficient and FL-friendly.

\section{Conclusion}
In this work, we have tackled adapting a foundational segmentation model (SAM) for 3D medical image segmentation by incorporating an effective PEFT strategy. We critically analyze the LoRA adapter's impact and various SAM components to make the fine-tuning for dense 3D segmentation tasks amenable to FL. Our approach simultaneously addresses data scarcity, overfitting, and communication overhead challenges, resulting in a practical and cost-efficient solution. Our current work analyses various fine-tuning methods in the context of FedAVG\cite{fed-fedaverage}; an interesting future direction would be studying the effects of various federated optimization strategies on low-rank adapters for datasets with considerable distribution shifts.

\begin{credits}
\subsubsection{\discintname}
The authors have no competing interests to declare that are relevant to the content of this article.
\end{credits}

%
%
\bibliographystyle{splncs04}
\bibliography{references}

\newpage
\appendix
\section{Appendix}
\renewcommand{\thetable}{A}
\begin{table}[h!]
\centering
\caption[ Results across local site test data on Prostate MRI dataset for different fine-tuning methods.]{Results on Prostate MRI dataset for different fine-tuning methods. 
$\ddag-$ FL setting of MA-SAM is not shown due to challenges in decomposing FacT tensors after federated aggregation.
The baseline (full fine-tuning) is highlighted in \colorbox{BaseColor}{Blue} and our method in \colorbox{OurColor}{Pink}. 
The best and second best methods for FL are highlighted in \textbf{bold} and \underline{underline}, respectively. 
Best viewed in color.
}
\label{tab:main_prostate}
{\renewcommand{\arraystretch}{1.2}
\resizebox{\columnwidth}{!}{
\begin{tabular}{crccccccc}
\hline
\multirow{2}{*}{\textbf{\begin{tabular}[c]{@{}c@{}}Experiments \\ \scriptsize{(Trainable/Total params)} \end{tabular}}} & \multirow{2}{*}{\textbf{Setting}} & \multicolumn{7}{c}{\textbf{Dice score}} \\ \cline{3-9} 
\multicolumn{1}{l}{} &  & \textbf{SiteA} & \textbf{SiteB} & \textbf{SiteC} & \textbf{SiteD} & \textbf{SiteE} & \textbf{SiteF} & \textbf{Average}\\ \hline

\rowcolor{BaseColor} \cellcolor{BaseColor} & Local & 0.9152 & 0.8551 & 0.8471 & 0.6781 & 0.9016 & 0.8556 & 0.8421 \\
\rowcolor{BaseColor} \cellcolor{BaseColor} & Federated & \textbf{0.9432} & \textbf{0.9109} & \textbf{0.8797} & \textbf{0.8984} & \textbf{0.9148} & \textbf{0.9036} & \textbf{0.9084} \\
\rowcolor{BaseColor} \multirow{-3}{*}{\begin{tabular}[c]{@{}c@{}}\textbf{FullFT (Baseline)}\\ \{$\theta_{\texttt{IE}}, \theta_{\texttt{PE}}, \theta_{\texttt{MD}}$\} \\ \scriptsize{(\textbf{90.399M}/90.399M) $\textbf{100\%}$} \end{tabular}}
 & Centralized & 0.9336 & 0.8926 & 0.8618 & 0.8851 & 0.9012 & 0.8989 & 0.8955 \\ \hline

\multirow{3}{*}{\begin{tabular}[c]{@{}c@{}} \textbf{AttnFT}\\ \{$\theta_{\texttt{IE-AT}}, \theta_{\texttt{MD}}$\} \\ \scriptsize{(\textbf{29.575M}/90.399M) $32.7\%$ } \end{tabular}}
 & Local & 0.5103 & 0.8953 & 0.2937 & 0.3754 & 0.8844 & 0.8301 & 0.6315 \\
 & Federated & 0.9243 & 0.8863 & 0.8516 & 0.8362 & 0.8960 & 0.8839 & 0.8797 \\
 & Centralized & 0.9312 & 0.8997 & 0.8637 & 0.8768 & 0.9087 & 0.8943 & 0.8957 \\ \hline

\multirow{3}{*}{\begin{tabular}[c]{@{}c@{}} \textbf{DecFT}\\ \{$\theta_{\texttt{MD}}$\} \\ \scriptsize{(\textbf{3.768M}/90.399M) $4.2\%$} \end{tabular}}
 & Local & 0.9123 & 0.8297 & 0.8005 & 0.7734 & 0.8121 & 0.7922 & 0.8200 \\
 & Federated & 0.9019 & 0.8427 & 0.7561 & 0.8097 & 0.7164 & 0.8337 & 0.8101 \\
 & Centralized & 0.9104 & 0.8549 & 0.8349 & 0.8348 & 0.8738 & 0.8433 & 0.8587 \\ \hline
 
\multirow{3}{*}{\begin{tabular}[c]{@{}c@{}} \textbf{LoRAFT}\\ \{$\theta_{\texttt{LoRA}}$\} \\ \scriptsize{(\textbf{1.368M}/91.767M) $1.5\%$} \end{tabular}}
 & Local & 0.8661 & 0.8856 & 0.8409 & 0.7769 & 0.8406 & 0.8218 & 0.8386 \\
 & Federated & 0.9045 & 0.8632 & 0.8283 & 0.8510 & 0.8595 & 0.8402 & 0.8578 \\
 & Centralized & 0.9229 & 0.8985 & 0.8541 & 0.8699 & 0.9083 & 0.8822 & 0.8893 \\ \hline

\multirow{3}{*}{\begin{tabular}[c]{@{}c@{}} \textbf{LoRADecFT}\\ \{$\theta_{\texttt{MD}}, \theta_{\texttt{LoRA}}$\} \\ \scriptsize{(\textbf{5.136M}/91.767M) $5.7\%$} \end{tabular}}
 & Local & 0.9355 & 0.9132 & 0.8639 & 0.8542 & 0.9063 & 0.8842 & 0.8929 \\
 & Federated & 0.9283 & \underline{0.8996} & \underline{0.8615} & \underline{0.8798} & \underline{0.9065} & \underline{0.8939} & \underline{0.8949} \\
 & Centralized & 0.9416 & 0.9106 & 0.8606 & 0.8906 & 0.9130 & 0.9072 & 0.9039 \\ \hline

\multirow{3}{*}{\begin{tabular}[c]{@{}c@{}} \textbf{PDecFT}\\ \{$\theta_{\texttt{UP}}, \theta_{\texttt{HYP}}$\} \\ \scriptsize{(\textbf{0.344M}/90.399M) $0.4\%$} \end{tabular}}
 & Local & 0.8719 & 0.7825 & 0.7969 & 0.7308 & 0.7687 & 0.7830 & 0.7890 \\
 & Federated & 0.8526 & 0.7790 & 0.7176 & 0.7066 & 0.4350 & 0.7197 & 0.7017 \\
 & Centralized & 0.8783 & 0.7893 & 0.7853 & 0.7769 & 0.7901 & 0.7848 & 0.8008 \\ \hline
 
\rowcolor{OurColor} & Local & 0.9339 & 0.9037 & 0.8634 & 0.8206 & 0.9063 & 0.8791 & 0.8845 \\
\rowcolor{OurColor} \cellcolor{OurColor} & Federated & \underline{0.9336} & 0.8886 & 0.8516 & 0.8665 & 0.8991 & 0.8808 & 0.8867 \\
\rowcolor{OurColor} \multirow{-3}{*}{\begin{tabular}[c]{@{}c@{}} \textbf{FLAP-SAM (Ours)}\\ \{$\theta_{\texttt{LoRA}}, \theta_{\texttt{UP}}, \theta_{\texttt{HYP}}$\} \\ \scriptsize{(\textbf{1.712M}/91.767M) $\textbf{1.9\%}$} \end{tabular}}
\cellcolor{OurColor} & Centralized & 0.9392 & 0.9081 & 0.8636 & 0.8992 & 0.9103 & 0.9058 & 0.9044 \\ \hline

{\begin{tabular}[c]{@{}c@{}} \textbf{MA-SAM$^\ddag$}\\ \scriptsize{(\textbf{28.667M}/115.297M) $25\%$} \end{tabular}}
 & Centralized & 0.945 & 0.922 & 0.877 & 0.897 & 0.923 & 0.910 & 0.912 \\ \hline

\end{tabular}}
}
\end{table}


\renewcommand{\thetable}{B}
\begin{table}[h!]
\centering
\caption[Results across local site test data on Fed-KiTS2019 dataset for different fine-tuning methods.]{Results on Fed-KiTS2019 dataset for different fine-tuning methods. 
$\ddag-$ FL setting of MA-SAM is not shown due to challenges in decomposing FacT tensors after federated aggregation.
The baseline (full fine-tuning) is highlighted in \colorbox{BaseColor}{Blue} and our method in \colorbox{OurColor}{Pink}. 
The best and second best methods for FL are highlighted in \textbf{bold} and \underline{underline}, respectively.
}
\label{tab:main_fed-kits}
{\renewcommand{\arraystretch}{1.2}
\resizebox{\columnwidth}{!}{
\begin{tabular}{crccccccc}
\hline
\multirow{2}{*}{\textbf{\begin{tabular}[c]{@{}c@{}}Experiments \\ \scriptsize{(Trainable/Total params)} \end{tabular}}} & \multirow{2}{*}{\textbf{Setting}} & \multicolumn{7}{c}{\textbf{Dice score}} \\ \cline{3-9} 
\multicolumn{1}{l}{} &  & \textbf{SiteA} & \textbf{SiteB} & \textbf{SiteC} & \textbf{SiteD} & \textbf{SiteE} & \textbf{SiteF} & \textbf{Average}\\ \hline

\rowcolor{BaseColor} \cellcolor{BaseColor} & Local & 0.4689 & 0.3830 & 0.4652 & 0.4566 & 0.4767 & 0.4454 & 0.4493 \\
\rowcolor{BaseColor} \cellcolor{BaseColor} & Federated & 0.6270 & 0.5258 & 0.5634 & 0.4800 & 0.5732 & 0.4967 & 0.5444 \\
\rowcolor{BaseColor} \multirow{-3}{*}{\begin{tabular}[c]{@{}c@{}}\textbf{FullFT (Baseline)}\\ \{$\theta_{\texttt{IE}}, \theta_{\texttt{PE}}, \theta_{\texttt{MD}}$\} \\ \scriptsize{(\textbf{90.533M}/90.533M) $\textbf{100\%}$} \end{tabular}}
 & Centralized & 0.6955 & 0.4273 & 0.5823 & 0.4705 & 0.5189 & 0.4698 & 0.5274 \\ \hline

\multirow{3}{*}{\begin{tabular}[c]{@{}c@{}} \textbf{AttnFT}\\ \{$\theta_{\texttt{IE-AT}}, \theta_{\texttt{MD}}$\} \\ \scriptsize{(\textbf{30.186M}/90.533M) $33.3\%$ } \end{tabular}}
 & Local & 0.5074 & 0.4306 & 0.5153 & 0.4673 & 0.5052 & 0.4772 & 0.4838 \\
 & Federated & 0.6749 & 0.4942 & 0.6025 & \underline{0.5534} & \underline{0.5839} & 0.5252 & 0.5724 \\
 & Centralized & 0.7015 & 0.4789 & 0.5972 & 0.4999 & 0.5164 & 0.4977 & 0.5486 \\ \hline

\multirow{3}{*}{\begin{tabular}[c]{@{}c@{}} \textbf{DecFT}\\ \{$\theta_{\texttt{MD}}$\} \\ \scriptsize{(\textbf{3.902M}/90.533M) $4.3\%$} \end{tabular}}
 & Local & 0.3754 & 0.3933 & 0.4285 & 0.4475 & 0.4037 & 0.4791 & 0.4213 \\
 & Federated & 0.5986 & 0.4404 & 0.5214 & 0.4627 & 0.5293 & 0.4882 & 0.5068 \\
 & Centralized & 0.6366 & 0.4288 & 0.5573 & 0.4732 & 0.5210 & 0.4902 & 0.5179 \\ \hline

\multirow{3}{*}{\begin{tabular}[c]{@{}c@{}} \textbf{LoRAFT}\\ \{$\theta_{\texttt{LoRA}}$\} \\ \scriptsize{(\textbf{1.368M}/91.901M) $1.5\%$} \end{tabular}}
 & Local & 0.3999 & 0.2885 & 0.3978 & 0.3816 & 0.3374 & 0.4247 & 0.3717 \\
 & Federated & 0.3984 & 0.2888 & 0.3697 & 0.3906 & 0.3470 & 0.4179 & 0.3687 \\
 & Centralized & 0.6028 & 0.4671 & 0.5228 & 0.4995 & 0.4981 & 0.5551 & 0.5242 \\ \hline

\multirow{3}{*}{\begin{tabular}[c]{@{}c@{}} \textbf{LoRADecFT}\\ \{$\theta_{\texttt{MD}}, \theta_{\texttt{LoRA}}$\} \\ \scriptsize{(\textbf{5.270M}/91.901M) $5.8\%$} \end{tabular}}
 & Local & 0.4800 & 0.5038 & 0.5607 & 0.4633 & 0.5177 & 0.5065 & 0.5053 \\
 & Federated & \underline{0.7193} & \underline{0.5172} & \underline{0.6155} & 0.5398 & \textbf{0.5861} & \textbf{0.6145} & \underline{0.5987} \\
 & Centralized & 0.7322 & 0.4908 & 0.6068 & 0.6291 & 0.5918 & 0.6095 & 0.6100 \\ \hline

\multirow{3}{*}{\begin{tabular}[c]{@{}c@{}} \textbf{PDecFT}\\ \{$\theta_{\texttt{UP}}, \theta_{\texttt{HYP}}$\} \\ \scriptsize{(\textbf{0.478M}/90.533M) $0.5\%$} \end{tabular}}
 & Local & 0.2992 & 0.3466 & 0.3994 & 0.3845 & 0.3696 & 0.4589 & 0.3764 \\
 & Federated & 0.5236 & 0.3660 & 0.4513 & 0.4307 & 0.4769 & 0.4730 & 0.4536 \\
 & Centralized & 0.5649 & 0.3992 & 0.4964 & 0.4516 & 0.4868 & 0.4766 & 0.4793 \\ \hline
 
\rowcolor{OurColor} & Local & 0.5107 & 0.4173 & 0.5738 & 0.4657 & 0.5289 & 0.5451 & 0.5069 \\
\rowcolor{OurColor} \cellcolor{OurColor} & Federated & \textbf{0.7234} & \textbf{0.5770} & \textbf{0.6169} &\textbf{0.5554} & 0.5762 & \underline{0.5786} & \textbf{0.6046} \\
\rowcolor{OurColor} \multirow{-3}{*}{\begin{tabular}[c]{@{}c@{}} \textbf{FLAP-SAM (Ours)}\\ \{$\theta_{\texttt{LoRA}}, \theta_{\texttt{UP}}, \theta_{\texttt{HYP}}$\} \\ \scriptsize{(\textbf{1.846M}/91.901M) $\textbf{2.0\%}$} \end{tabular}}
\cellcolor{OurColor} & Centralized & 0.7041 & 0.4500 & 0.6292 & 0.5772 & 0.6201 & 0.6075 & 0.5980 \\ \hline

{\begin{tabular}[c]{@{}c@{}} \textbf{MA-SAM$^\ddag$}\\ \scriptsize{(\textbf{28.721M}/116.605M) $25\%$} \end{tabular}}
 & Centralized & 0.725 & 0.544 & 0.605 & 0.513 & 0.566 & 0.661 & 0.602 \\ \hline

\end{tabular}}
}
\end{table}

\renewcommand{\thetable}{C}
\begin{table}[h!]
\centering
\caption[Results across local site test data on Fed-IXI dataset for different fine-tuning methods.]{Results on Fed-IXI dataset for different fine-tuning methods. 
$\ddag-$ FL setting of MA-SAM is not shown due to challenges in decomposing FacT tensors after federated aggregation.
The baseline (full fine-tuning) is highlighted in \colorbox{BaseColor}{Blue} and our method in \colorbox{OurColor}{Pink}. 
The best and second best methods for FL are highlighted in \textbf{bold} and \underline{underline}, respectively.
}
\label{tab:main_fed-ixi}
{\renewcommand{\arraystretch}{1.2}
\resizebox{\columnwidth}{!}{
\begin{tabular}{crcccc}
\hline
\multirow{2}{*}{\textbf{\begin{tabular}[c]{@{}c@{}}Experiments \\ \scriptsize{(Trainable/Total params)} \end{tabular}}} & \multirow{2}{*}{\textbf{Setting}} & \multicolumn{4}{c}{\textbf{Dice score}} \\ \cline{3-6} 
\multicolumn{1}{l}{} &  & \textbf{SiteA} & \textbf{SiteB} & \textbf{SiteC} & \textbf{Average}\\ \hline

\rowcolor{BaseColor} \cellcolor{BaseColor} & Local & 0.9808 & 0.9797 & 0.9728 & 0.9777 \\
\rowcolor{BaseColor} \cellcolor{BaseColor} & Federated & 0.9837 & 0.9835 & 0.9761 & 0.9811 \\
\rowcolor{BaseColor} \multirow{-3}{*}{\begin{tabular}[c]{@{}c@{}}\textbf{FullFT (Baseline)}\\ \{$\theta_{\texttt{IE}}, \theta_{\texttt{PE}}, \theta_{\texttt{MD}}$\} \\ \scriptsize{(\textbf{90.399M}/90.399M) $\textbf{100\%}$} \end{tabular}}
 & Centralized & 0.9838 & 0.9846 & 0.9818 & 0.9834 \\ \hline

\multirow{3}{*}{\begin{tabular}[c]{@{}c@{}} \textbf{AttnFT}\\ \{$\theta_{\texttt{IE-AT}}, \theta_{\texttt{MD}}$\} \\ \scriptsize{(\textbf{29.575M}/90.399M) $32.7\%$ } \end{tabular}}
 & Local & 0.9592 & 0.9708 & 0.7244 & 0.8848 \\
 & Federated & 0.9713 & 0.9701 & 0.9608 & 0.9674 \\
 & Centralized & 0.9781 & 0.9791 & 0.9749 & 0.9774\\ \hline

\multirow{3}{*}{\begin{tabular}[c]{@{}c@{}} \textbf{DecFT}\\ \{$\theta_{\texttt{MD}}$\} \\ \scriptsize{(\textbf{3.768M}/90.399M) $4.2\%$} \end{tabular}}
 & Local & 0.9783 & 0.9787 & 0.9680 & 0.9750 \\
 & Federated & 0.9780 & 0.9787 & 0.9746 & 0.9771 \\
 & Centralized & 0.9791 & 0.9806 & 0.9770 & 0.9789 \\ \hline

\multirow{3}{*}{\begin{tabular}[c]{@{}c@{}} \textbf{LoRAFT}\\ \{$\theta_{\texttt{LoRA}}$\} \\ \scriptsize{(\textbf{1.368M}/91.767M) $1.5\%$} \end{tabular}}
 & Local & 0.9788 & 0.9776 & 0.9620 & 0.9728 \\
 & Federated & 0.9790 & 0.9791 & 0.9749 & 0.9777 \\
 & Centralized & 0.9803 & 0.9808 & 0.9785 & 0.9798 \\ \hline

\multirow{3}{*}{\begin{tabular}[c]{@{}c@{}} \textbf{LoRADecFT}\\ \{$\theta_{\texttt{MD}}, \theta_{\texttt{LoRA}}$\} \\ \scriptsize{(\textbf{5.136M}/91.767M) $5.7\%$} \end{tabular}}
 & Local & 0.9850 & 0.9853 & 0.9784 & 0.9829 \\
 & Federated & \textbf{0.9850} & \textbf{0.9853} & \textbf{0.9806} & \textbf{0.9836} \\
 & Centralized & 0.9855 & 0.9864 & 0.9837 & 0.9852 \\ \hline

\multirow{3}{*}{\begin{tabular}[c]{@{}c@{}} \textbf{PDecFT}\\ \{$\theta_{\texttt{UP}}, \theta_{\texttt{HYP}}$\} \\ \scriptsize{(\textbf{0.344M}/90.399M) $0.4\%$} \end{tabular}}
 & Local & 0.9713 & 0.9722 & 0.9600 & 0.9678 \\
 & Federated & 0.9697 & 0.9725 & 0.9658 & 0.9693 \\
 & Centralized & 0.9718 & 0.9729 & 0.9685 & 0.9711 \\ \hline

\rowcolor{OurColor} & Local & 0.9850 & 0.9849 & 0.9788 & 0.9829 \\
\rowcolor{OurColor} \cellcolor{OurColor} & Federated & \underline{0.9847} & \underline{0.9852} & \underline{0.9803} & \underline{0.9834} \\
\rowcolor{OurColor} \multirow{-3}{*}{\begin{tabular}[c]{@{}c@{}} \textbf{FLAP-SAM (Ours)}\\ \{$\theta_{\texttt{LoRA}}, \theta_{\texttt{UP}}, \theta_{\texttt{HYP}}$\} \\ \scriptsize{(\textbf{1.712M}/91.767M) $\textbf{1.9\%}$} \end{tabular}}
\cellcolor{OurColor} & Centralized & 0.9854 & 0.9864 & 0.9835 & 0.9851 \\ \hline

{\begin{tabular}[c]{@{}c@{}} \textbf{MA-SAM$^\ddag$}\\ \scriptsize{(\textbf{28.667M}/115.297M) $25\%$} \end{tabular}}
 & Centralized & 0.972 & 0.972 & 0.968 & 0.971 \\ \hline

\end{tabular}}
}
\end{table}

\renewcommand{\thetable}{D}
\begin{table}[h!]
\centering
\caption[Dice score for different LoRA rank values]{Results across local site test data for different rank values of LoRA in the federated setting on Fed-KiTS19 dataset.}
\label{tab:ablation1}
{\renewcommand{\arraystretch}{1.2}
\resizebox{\columnwidth}{!}{
\begin{tabular}{ccccccccc}
\hline
\multirow{2}{*}{\textbf{\begin{tabular}[c]{@{}c@{}}LoRA \\ rank\end{tabular}}} & \multicolumn{7}{c}{\textbf{Dice score}} & \multirow{2}{*}{\textbf{\begin{tabular}[c]{@{}c@{}}Trainable /\\ Total params\end{tabular}}} \\ \cline{2-8}
& {\textbf{SiteA}} & {\textbf{SiteB}} & {\textbf{SiteC}} & {\textbf{SiteD}} & {\textbf{SiteE}} & {\textbf{SiteF}}  & \textbf{Average} & \\ \hline
\textbf{32} & 0.723 & 0.577 & 0.617 & 0.555 & 0.576 & 0.579 & 0.605 & 1.846M / 91.901M \\
\textbf{16} & 0.720 & 0.572 & 0.616 & 0.518 & 0.561 & 0.604 & 0.599 & 1.162M / 91.217M \\
\textbf{4}& 0.703 & 0.590 & 0.609 & 0.538 & 0.588 & 0.572 &  0.600 & 0.649M / 90.704M \\ \hline
\end{tabular}}
}
\end{table}
\renewcommand{\thetable}{E}
\begin{table}[h!]
\centering
\caption[Dice score for different PEFT methods]{Results across local test data site for different PEFT methods in the federated setting on Fed-KiTS19 dataset.}
\label{tab:ablation2}
{\renewcommand{\arraystretch}{1.2}
\resizebox{\columnwidth}{!}{
\begin{tabular}{ccccccccc}
\hline
\multirow{2}{*}{\textbf{Methods}} & \multicolumn{7}{c}{\textbf{Dice score}} & \multirow{2}{*}{\textbf{\begin{tabular}[c]{@{}c@{}}Trainable /\\ Total params\end{tabular}}} \\ \cline{2-8}
& {\textbf{SiteA}} & {\textbf{SiteB}} & {\textbf{SiteC}} & {\textbf{SiteD}} & {\textbf{SiteE}} & {\textbf{SiteF}}  & \textbf{Average} & \\ \hline
\textbf{LoRA}& 0.723 & \textbf{0.577} & \textbf{0.617} & \textbf{0.555} & 0.576 & 0.578 & \textbf{0.605} & 1.846M / 91.901M \\
\textbf{DoRA} & \textbf{0.725} & 0.470 & 0.615 & 0.537 & 0.609 & 0.594 & 0.592 & 1.846M / 91.901M \\
\textbf{MoLE} & 0.712 & 0.517 & 0.613 & 0.552 & \textbf{0.615} & \textbf{0.611} & 0.603 & 4.583M / 94.637M \\ \hline
\end{tabular}}
}
\end{table}

\end{document}